\pdfoutput=1

\documentclass[11pt]{article}

\usepackage[]{ACL2023}

\usepackage{times}
\usepackage{latexsym}
\usepackage{graphicx}

\usepackage{amsmath}
\DeclareMathOperator*{\argmax}{arg\,max}

\newcommand{\xml}[1]{$<${\fontfamily{cmtt}\selectfont {#1}}$>$}
\usepackage[T1]{fontenc}
\usepackage[utf8]{inputenc}

\usepackage{microtype}

\usepackage{inconsolata}

\title{Multi-Document Summarization with Centroid-Based Pretraining}

\author{Ratish Puduppully$^{1,2}$\thanks{~~Part of the work was done when the author was at the University of Edinburgh}\hspace{0.2cm} \textnormal{and} Parag Jain$^4$ \textnormal{and} Nancy F. Chen$^{1,2,3}$\textnormal{and} Mark Steedman$^4$\\
$^1$Institute for Infocomm Research (I$^2$R), A$^{*}$STAR, Singapore\\
$^2$CNRS@CREATE, Singapore\\
$^3$Centre for Frontier AI Research (CFAR), A*STAR\\
    $^4$School of Informatics, University of Edinburgh \\    
\texttt{puduppully@i2r.a-star.edu.sg}~~~~\texttt{parag.jain@ed.ac.uk}~~~~ \\\texttt{nfychen@i2r.a-star.edu.sg}~~~~\texttt{steedman@inf.ed.ac.uk}\\
}

\begin{document}
\maketitle
\begin{abstract}
In Multi-Document Summarization (MDS), the input can be modeled as a set of documents, and the output is its summary. In this paper, we focus on pretraining objectives for MDS. Specifically, we introduce a novel pretraining objective, which involves selecting the ROUGE-based centroid of each document cluster as a proxy for its summary. Our objective thus does not require human written summaries and can be utilized for pretraining on a dataset consisting solely of document sets. Through zero-shot, few-shot, and fully supervised experiments on multiple MDS datasets, we show that our model \textit{Centrum} is better or comparable to a state-of-the-art model. We make the pretrained and fine-tuned models freely available to the research community\footnote{\url{https://github.com/ratishsp/centrum}}.

\end{abstract}

\section{Introduction}
In Multi-Document Summarization (MDS), the input is a set of documents, and the 
output is a summary that describes important information in a 
coherent and non-redundant manner \cite{10.1145/215206.215334,radev-mckeown-1998-generating}.
In recent years, there have been significant improvements in MDS due to the availability of 
MDS datasets \cite{fabbri-etal-2019-multi,gholipour-ghalandari-etal-2020-large,j.2018generating}
and advances in pretraining approaches \cite{lewis-etal-2020-bart,DBLP:journals/jmlr/RaffelSRLNMZLL20,DBLP:conf/icml/ZhangZSL20}.

In particular, \citet{xiao-etal-2022-primera} introduced a pretraining approach called PRIMERA (Pyramid-based Masked Sentence Pretraining) adapted for MDS.
To create synthetic summaries, they used the Pyramid scheme \cite{nenkova-passonneau-2004-evaluating}, incorporating a process of identifying and ranking entities, followed by grouping sentences containing these entities in the input documents. The sentences with the highest overlap with other documents (measured using ROUGE) in each group were masked in the input and integrated into the output, forming a synthetic summary. 
\citet{xiao-etal-2022-primera} show that an encoder-decoder model trained on such a corpus attains strong zero-shot, few-shot, and fully supervised results on multiple datasets. 

However, these synthetic summaries may lack coherence as the sentences are derived from various positions within the input documents.
Furthermore, there is potential for redundancy, as sentences encapsulating similar information could be selected for inclusion in the summary.

In this paper, we propose Centrum, a pretraining objective for MDS, which is conceptually simple and overcomes these problems. 
The key intuition is that among a set of documents in a news cluster, the document which shares the most content with the other documents in the cluster can serve as a proxy for the summary of the document set.
Such a cluster centroid is inherently coherent as it is a human-written document. 
Furthermore, because it isn't artificially assembled, it avoids content repetition.

In this paper, we pretrain Centrum on NewSHead \cite{headline2020} corpus and perform zero-shot, few-shot and fully-supervised experiments on various MDS datasets. 
We show that Centrum performs favorably compared to PRIMERA, especially in the zero-shot and few-shot settings, where there are none or very few training examples available.

\section{Centroid-based Selection of Document as Summary}
\label{sec:select}
\paragraph{Background on PRIMERA}
\citet{xiao-etal-2022-primera} leveraged the NewSHead corpus \cite{headline2020}, a compilation of 369,940 news clusters, for pretraining.
Using the Pyramid scheme \cite{nenkova-passonneau-2004-evaluating}, they created synthetic summaries through a multi-step procedure.
They gathered the entity mentions in the input documents and rank the entities by the count of documents in which an entity is mentioned.
Next, they divide the sentences from the documents into distinct groups, such that the sentences containing an entity belong to the same group.
They then extracted the sentence with the highest overlap (as quantified by ROUGE \cite{lin-2004-rouge}) with other documents from each group.
This sentence was replaced with a mask token in the input, and copied to the output document.
The idea here was to leverage information from other documents to reconstruct the masked sentence. The sentences thus obtained were concatenated to form a synthetic summary. 

\citet{xiao-etal-2022-primera} applied the method to the  NewSHead corpus \cite{headline2020} containing news articles clustered by topic. 
To accommodate long document lengths, they use Longformer Encoder-Decoder (LED) architecture \cite{beltagy2020longformer}. 
LED supports sparse global attention along with dense local attention on the input. 
PRIMERA employs global attention on specialized tokens (\xml{doc-sep}), which act as separators between the documents within the input cluster. 
The pretrained PRIMERA model was then used for zero-shot evaluation, few-shot or full finetuning across multiple MDS datasets.

\paragraph{Problems with PRIMERA pretraining}
PRIMERA's reference summaries consist of sentences extracted from varying positions within different documents in a cluster. This method can yield incoherent summaries, as it can be unclear which entities the sentences refer to.
We illustrate this with an example of a synthetic summary created using PRIMERA in Table~\ref{tab:synthetic-summary}.
The first sentence about Lady Gaga originates from the first document, while the second sentence mentioning Donald Trump and Elton John comes from the second document. 
The lack of entity mentions within these sentences disrupts the overall coherence of the summary.
We also note occurrences of redundant information in the synthetic summary.
Our hypothesis is that pretraining using such noisy synthetic summaries could negatively impact model performance, particularly in zero-shot or few-shot experiments.

\begin{table}[t]
    \centering
    \small
    \begin{tabular}{p{\columnwidth}}
         \textcolor{orange}{She's a fantastic person, solid as a rock and I'm very proud of her success because I really believe I had at least something to do with it." It was unclear exactly what type of records he was referring to — the attendance of 6,500 fell far short of many Elton John concerts.} $\dots$ (4 sent) Donald Trump made his way to Great Falls, Montana, on Thursday (July 5), primarily to slam Democratic Sen. Jon Tester and accuse him of failing to live up to his promises in Washington. \textcolor{red}{``I’ve broken more Elton John [attendance] records, and I don’t have a musical instrument,'' he boasted.} \textcolor{brown}{This is my only musical instrument-the mouth-and hopefully the brain is attached to the mouth.} During a rally in Great Falls, Montana, where President Trump derided the \#MeToo movement and attacked individual Democratic lawmakers, the president once again bragged about the size of his supporter turnout. \textcolor{red}{``I’ve broken more Elton John [attendance] records, and I don’t have a musical instrument,'' Trump said according to Yahoo News.} ``I don’t have a guitar, or an organ. $\dots$ \textcolor{brown}{This is my only musical instrument - the mouth - and hopefully the brain is attached to the mouth.} The brain is so much more important.'' $\dots$\\
    \end{tabular}
    \caption{Example of a synthetic reference summary in PRIMERA \cite{xiao-etal-2022-primera}. We see that the reference summaries in PRIMERA can contain instances of incoherence and repetition. In this summary, the first sentence is about Lady Gaga, and the second is about Donald Trump and Elton John. The subjects of the first two sentences (highlighted in \textcolor{orange}{orange}) are unclear due to the lack of named entities. Additionally, the sentences in \textcolor{brown}{brown} and \textcolor{red}{red} contain repetitive information. }
    \label{tab:synthetic-summary}
\end{table}

\paragraph{Our Model}
We propose an alternate pretraining objective for MDS called as Centrum. 
We hypothesize that a document exhibiting the highest similarity with the rest of the documents in a cluster could serve as a proxy for its summary.
This method inherently filters out documents that bear only a distant relation to other documents in the cluster. 
Furthermore, it addresses potential noise present in automatically created multi-document cluster datasets \cite{headline2020}, for example, a document falsely associated with a cluster.
The Centrum pretraining objective excludes such noise, as a mismatched document would not be chosen as the cluster centroid.
Among the documents, a document may have more relevant content than others. The Centrum objective will select the more relevant document as the summary.

Drawing inspiration from \citet{headline2020}, we designate a document as the summary if it maximizes the semantic match with other documents in the cluster. Specifically, from each document set $\mathcal{D}$ in an instance, the best candidate summary $\hat{y}$ is selected as:
\begin{equation}
    \hat{y} = \argmax_{x \in \mathcal{D}} 1/|\mathcal{D}| \sum_{x' \in D\setminus\{x\}} f(x, x') \label{eqn:best-document}
\end{equation}
where $f(x, x')$ represents the semantic match of summary $x$ with document $x'$. A model can be trained to learn this function $f$. In our approach, we employ the average of ROUGE1, ROUGE2, and ROUGEL as this function. Our pretraining corpus is constructed by treating $D\setminus\{\hat{y}\}$ as the input and $\hat{y}$ as the output.

\citet{DBLP:journals/corr/abs-2201-02321} recently employed a comparable strategy for unsupervised MDS. However, our approach differs from theirs by applying this strategy for MDS task-specific pretraining. Moreover, following \citet{xiao-etal-2022-primera}, we employ the LED architecture for handling long document context in the input.

\section{Experimental Setup}
\paragraph{Model}
We utilize the Transformers \cite{wolf-etal-2020-transformers} library to conduct our experiments. 
Similar to \citet{xiao-etal-2022-primera}, we adopt the large configuration of LED, comprising 459M parameters.
We finetune the LED model on the NewSHead corpus \cite{headline2020} with our Centrum pretraining objective. 
Documents within a cluster are concatenated into a single text sequence, with \xml{doc-sep} tokens employed as separators.
We apply global attention to the \xml{doc-sep} tokens, while local attention is used for the remaining tokens.
Further details about the hyperparameter settings can be found in Appendix \ref{sec:additional-hyperparams}.

\paragraph{Datasets} 
We conduct our evaluation on the Multi-News \cite{fabbri-etal-2019-multi}, WCEP \cite{gholipour-ghalandari-etal-2020-large}, and DUC 2007 datasets, comparing zero-shot, few-shot, and fully-supervised results. DUC 2007 comprises 45 examples, 20 of which we designate as the test set \cite{xiao-etal-2022-primera}.

\paragraph{Preprocessing of NewSHead Dataset}
We apply the following criteria when preprocessing the dataset:
\begin{itemize}
\item \textbf{Minimum Document Count in a Cluster}: We require that a news cluster must contain a minimum of three documents, allowing a document to serve as a summary for the remaining documents in the cluster. Clusters not meeting this requirement are excluded.
\item \textbf{Minimum Summary Size}: We hypothesize that a significant variance in summary lengths during pretraining could hurt performance. Therefore, we ensure that candidate summaries during pretraining are not too short, setting a minimum requirement of 250 tokens. Clusters not meeting this requirement are also excluded. In contrast, \citet{xiao-etal-2022-primera} can control the length of their synthetic reference summaries, ensuring that the sentence count in the synthetic summary constitutes at least 30\% of the total sentences in the cluster.
\end{itemize}
Additional preprocessing steps are outlined in Appendix \ref{sec:additional-preprocessing}. After applying these criteria, we retain 172K clusters, approximately 45\% of the total clusters in the NewSHead corpus \cite{headline2020}.

\paragraph{Comparison models}
In addition to the reported scores of PRIMERA (denoted as PRIMERA* in Table \ref{tab:scores4}), we independently reproduce the PRIMERA model scores by running inference using the PRIMERA checkpoints available in the Transformers \cite{wolf-etal-2020-transformers} library. Similar to the findings of \citet{giorgi-et-al}, we note that our reproduced scores are lower than those reported by \citet{xiao-etal-2022-primera}, an exception being the zero-shot results for the WCEP dataset. 
For a broader comparison, we also consider the Pegasus model proposed by \citet{DBLP:conf/icml/ZhangZSL20}.
Pegasus is a pretrained model focusing on single-document summarization (SDS), which obtains 
strong results on multiple SDS datasets such as XSum \cite{narayan-etal-2018-dont} and 
CNN-DailyMail \cite{DBLP:conf/nips/HermannKGEKSB15}.

\begin{table*}[t]
    \centering
    \small
    \begin{tabular}{p{3cm}|ccc|ccc|ccc}%
   & \multicolumn{3}{c|}{Zero Shot} &\multicolumn{3}{c|}{10 Examples} &\multicolumn{3}{c}{100 Examples} \\\hline
System &  R1 & R2 & RL&R1 & R2 & RL & R1 & R2 & RL\\\hline    
\multicolumn{10}{c}{Multi-News (256)} \\     \hline  
         Pegasus* & 32.0 & 10.1 & 16.7 & 39.0&12.1&20.3 &  43.0&13.5&21.1 \\ 
         PRIMERA* &42.0 &13.6 &20.8 & 44.0&15.5&22.0 &46.0&16.8&22.9\\ 
         PRIMERA &41.6 & 13.1 & 19.9 & 43.4 & 15.3 & 21.6 &45.2 &16.3 & 22.7\\
         Centrum & \textbf{43.5} &\textbf{15.7} & \textbf{22.4} & \textbf{43.4}  & \textbf{16.6} & \textbf{22.2} & \textbf{45.7} & \textbf{16.8} & \textbf{23.2} \\ \hline
 \multicolumn{10}{c}{WCEP (50)} \\    \hline
Pegasus* & 33.2 & 12.7 &23.8 & 35.6&14.8&26.8 &42.1&19.9&33.0 \\ 
PRIMERA* & 28.0 & 10.3 & 20.9 & 39.0&17.6&30.6 &43.0&20.5&33.9 \\     
PRIMERA &32.9&12.1& 23.4 &37.0&15.8&28.2&\textbf{42.4}&\textbf{20.5}&\textbf{33.4}\\
Centrum & \textbf{35.7} &\textbf{14.2} & \textbf{25.8} & \textbf{38.2} & \textbf{17.0} & \textbf{29.5} & 42.0 & 20.1 & 33.0 \\\hline 
\multicolumn{10}{c}{DUC2007 (250)} \\    \hline
Pegasus &22.7 & 4.2 & 12.8 &23.1 &3.5 &15.2 &- &- &- \\
PRIMERA &31.9 & 5.4 & 14.2 & 34.6 & 6.6 & 15.2 &- &- &- \\
Centrum &\textbf{32.7} & \textbf{5.7} & \textbf{15.0} & \textbf{35.3} & \textbf{7.7} & \textbf{16.8} & -& -& -\\ \hline
    \end{tabular}
    \caption{This table presents the ROUGE scores for zero-shot and few-shot evaluations on the Multi-News, WCEP, and DUC datasets. PRIMERA* represents the scores reported by \citet{xiao-etal-2022-primera}, while PRIMERA corresponds to the scores we reproduced using their provided checkpoints. The figures in parentheses denote the maximum length set during inference. Due to the DUC 2007 dataset's total size of 45 examples, results for few-shot evaluations with 100 examples are not provided. Our proposed model, Centrum, surpasses PRIMERA in zero-shot and few-shot (10 examples) settings, and performs comparably in the few-shot (100 examples) setting.}
    \label{tab:scores4}
\end{table*}
\section{Results}
We conduct experiments in three settings: zero-shot, few-shot, and fully supervised. 

\paragraph{Zero-shot}  
In the zero-shot setting, we evaluate our pretrained Centrum model on the test datasets of Multi-News, WCEP, and DUC 2007. Following \citet{xiao-etal-2022-primera}, the output length of the summary is set as the average length of the gold summaries of the test datasets. As Table \ref{tab:scores4} illustrates, Centrum outperforms the PRIMERA model in terms of ROUGE scores across all three datasets.

\paragraph{Few-shot} In the few-shot setting, we follow the approach of \citet{xiao-etal-2022-primera} by conducting two sets of experiments. We randomly select 10 and 100 examples from the training set for model finetuning, and an equivalent number of examples from the validation set. 
To account for potential variance in scores due to example selection, we repeat this process five times with different seeds.

We observe that the summaries generated by Centrum are, on average, longer than those produced by PRIMERA. This is primarily a result of the Centrum pretraining objective, which imposes a minimum summary length of 250 tokens. In contrast, PRIMERA synthetic summaries are restricted to a maximum length equating to 30\% of the input set. To ensure a fair comparison, we truncate the summaries in the few-shot setting to match the lengths assigned in the zero-shot setting.

Table \ref{tab:scores4} presents the average scores obtained over the five seeds. Given that the DUC 2007 dataset contains only 45 examples, results are reported for training and validation with 10 examples. From the results, we see that Centrum outperforms PRIMERA across all datasets when finetuned with 10 examples. Furthermore, Centrum maintains performance parity with PRIMERA when finetuned using 100 examples.

\begin{table}[t]
    \centering
    \small
    \begin{tabular}{p{3cm}|c|c|c}
         System &  R1 & R2 & RL\\\hline
         PRIMERA* &49.9&21.1&25.9\\ 
         PRIMERA & \textbf{50.0}&\textbf{20.6}&\textbf{25.5}\\
         Centrum&49.0 &20.4 & 25.4 \\\hline
    \end{tabular}
    \caption{Comparison of fully-supervised models based on ROUGE scores on the Multi-News dataset. Our proposed model, Centrum, demonstrates performance on par with PRIMERA.}
    \label{tab:scores3}
\end{table}

\paragraph{Fully supervised} In this setting, the pretrained models are finetuned on the training split of the Multi-News dataset. As reported in Table \ref{tab:scores3}, the results from the fully-supervised experiments demonstrate that Centrum performs on par with PRIMERA on the Multi-News dataset.

\paragraph{Human Evaluation}
To complement the automatic evaluation results, we conduct a human evaluation study. Three professional linguists are tasked with comparing the outputs of Centrum, PRIMERA, and Pegasus using the DUC 2007 dataset, and are compensated at rates higher than local minimum wages.
The evaluation focuses on three metrics as outlined by \citet{angelidis-etal-2021-extractive}: informativeness (which assesses the consistency between model output and the human reference summary), coherence (which evaluates the ordering of information in the summary), and non-repetition (where a higher-quality summary exhibits fewer repetitions of words, phrases, or sentences). 

The evaluators are presented with three summaries from the three models, randomly ordered, along with the reference summary. They are then instructed to rank the summaries from best (+1) to worst (-1) for each of the three metrics. These rankings are summed and scaled by the number of examples (20), producing scores that range from 100\% (best) to --100\% (worst). The results of this human evaluation are presented in Table \ref{tab:human_eval}. 

Our findings indicate that Centrum significantly outperforms Pegasus across all three metrics, as confirmed by a one-way ANOVA with a post-hoc Tukey test (p $\leq$ 0.05). In comparison to PRIMERA, Centrum is significantly better in terms of informativeness and performs comparably in terms of coherence and non-repetition. Pegasus, on the other hand, is marked by heavy repetition within its summaries, which likely accounts for its lower scores.

\begin{table}[t]
    \centering
    \small
    \begin{tabular}{c|c|c|c}
         &  Inform & Coh & Rep\\\hline
         Pegasus&-100* & -100*& -100*\\\hline
         PRIMERA&34.5* & 46.6& 58.6 \\ \hline
         Centrum&65.5 & 53.4& 41.4 \\ \hline
    \end{tabular}
\caption{Human evaluation results for the DUC2007 dataset, with higher scores being preferable. We compare the Pegasus, PRIMERA, and Centrum models across three metrics: informativeness (Inform), coherence (Coh), and avoidance of repetition (Rep). Results that are statistically significantly different from Centrum are marked with an asterisk (*).}
    \label{tab:human_eval}
\end{table}

\section{Conclusion}
We propose a centroid-based pretraining objective for multi-document summarization. 
Through experiments, we see that 
our model Centrum outperforms the existing state-of-the-art model PRIMERA on zero-shot settings and 
is comparable with PRIMERA in few-shot and supervised settings. 

\section{Limitations}
As mentioned in the main paper, one of the limitations of our Centrum model is that it tends to produce longer outputs in comparison to PRIMERA. 
This necessitates controlling the length of the summary by truncating to a desired length. 
Moreover, due to our requirement of at least three documents in a cluster for centroid computation, we are unable to utilize clusters of only two documents present in \citet{headline2020}. 
This constraint significantly reduces the utilizable corpus size, leading us to work with roughly 45\% of the corpus size used by PRIMERA.
Future research could explore the possibility of initializing Centrum with the gap sentence generation-based Pegasus \cite{DBLP:conf/icml/ZhangZSL20} single document summarization objective, potentially allowing for full utilization of the corpus size of \citet{headline2020}.

\section*{Acknowledgements}
This research was supported by funding from the Institute for Infocomm Research (I2R) under A*STAR ARES, Singapore, and by the National Research Foundation, Prime Minister’s Office, Singapore under its Campus for Research Excellence and Technological Enterprise (CREATE) programme.
The work was supported in part by ERC Advanced
Fellowship GA 742137 SEMANTAX and the University of Edinburgh Huawei
Laboratory. Parag is supported by Huawei and the UKRI Centre for Doctoral Training in Natural Language Processing (grant EP/S022481/1).
We thank the anonymous reviewers for their constructive feedback.

\bibliography{anthology,custom}
\bibliographystyle{acl_natbib}

\appendix
\section{Potential Risks}
Despite our model's potential, there is a risk that the generated summaries might not accurately represent the input document due to noise present in the training and finetuning examples. At the same time, we believe that our Centrum pretraining strategy doesn't affect the factuality of the model either positively or negatively compared to \citet{xiao-etal-2022-primera}. 
Future research will aim to explicitly evaluate and improve the factuality of our model's output.

\section{Details of the Datasets}
\begin{table}[t]
    \centering
    \small
    \begin{tabular}{c|c|c|c|c}
         Name& \#Ex & \#Doc/C & \#L$_{doc}$ & \#L$_{summ}$  \\ \hline
         NewSHead (\citeyear{headline2020})& 177K& 4.2 & 1692& 484\\\hline
         Multi-News (\citeyear{fabbri-etal-2019-multi}) & 56K &2.8 & 1793 & 217 \\
         WCEP (\citeyear{gholipour-ghalandari-etal-2020-large})  &10K & 9.1&3866 &28 \\
         DUC 2007 &45 & 25&540 &250\\\hline
    \end{tabular}
\caption{Characteristics of the datasets utilized in this paper. The notations are as follows: \#Ex represents the number of examples, \#Doc/C is the average number of documents per cluster, \#L$_{doc}$ signifies the average token count in the input, and \#L$_{summ}$ indicates the average token count in the summary. Values associated with the Multi-News and WCEP datasets are sourced from \citet{xiao-etal-2022-primera}.}
    \label{tab:datasets}
\end{table}
Table \ref{tab:datasets} provides detailed information about the datasets used in our study. The NewSHead, Multi-News, and DUC 2007 datasets all originate from the news domain, while the WCEP dataset is derived from the Wikipedia Current Events Portal.

\section{Hyperparameter Details}
\label{sec:additional-hyperparams}
Our hyperparameters are similar to \citet{xiao-etal-2022-primera}.
We train for 100K steps with a learning rate of 3e-5. We evaluate every 500 steps and early-stop on the validation perplexity with a patience of 50.
Pretraining Centrum on a 4-node A100 GPU took around 4 days.
We computed the results using ROUGE \cite{lin-2004-rouge} library \footnote{\url{https://github.com/google-research/google-research/tree/master/rouge}} with the default settings and `--use\_stemmer' argument.

\section{Additional Preprocessing Steps}
\label{sec:additional-preprocessing}
\begin{itemize}
\item \textbf{Removing boilerplate text from summaries}: We remove boilerplate text such as ``Sorry, this video isn't available any more.'', 
``Advertisement Story continues below'' from the summary sentences using regular 
expression based cleaning.
\item \textbf{Truncation of documents}: We truncate each document in the cluster to 
the maximum length of source context allowed in LED divided by the count of the documents in the cluster. 
Thus, each document
has a proportional representation in the cluster, similar to \citet{xiao-etal-2022-primera}.
\end{itemize}

\section{Software and Licenses}
\label{sec:software-and-licenses}
Our model relies on datasets downloaded from HuggingFace datasets \cite{lhoest-etal-2021-datasets} (Apache 2.0). 
We release our models under the Apache 2.0 license.

\section{Human Evaluation}
\begin{figure*}
    \centering
    \includegraphics[width=0.9\textwidth]{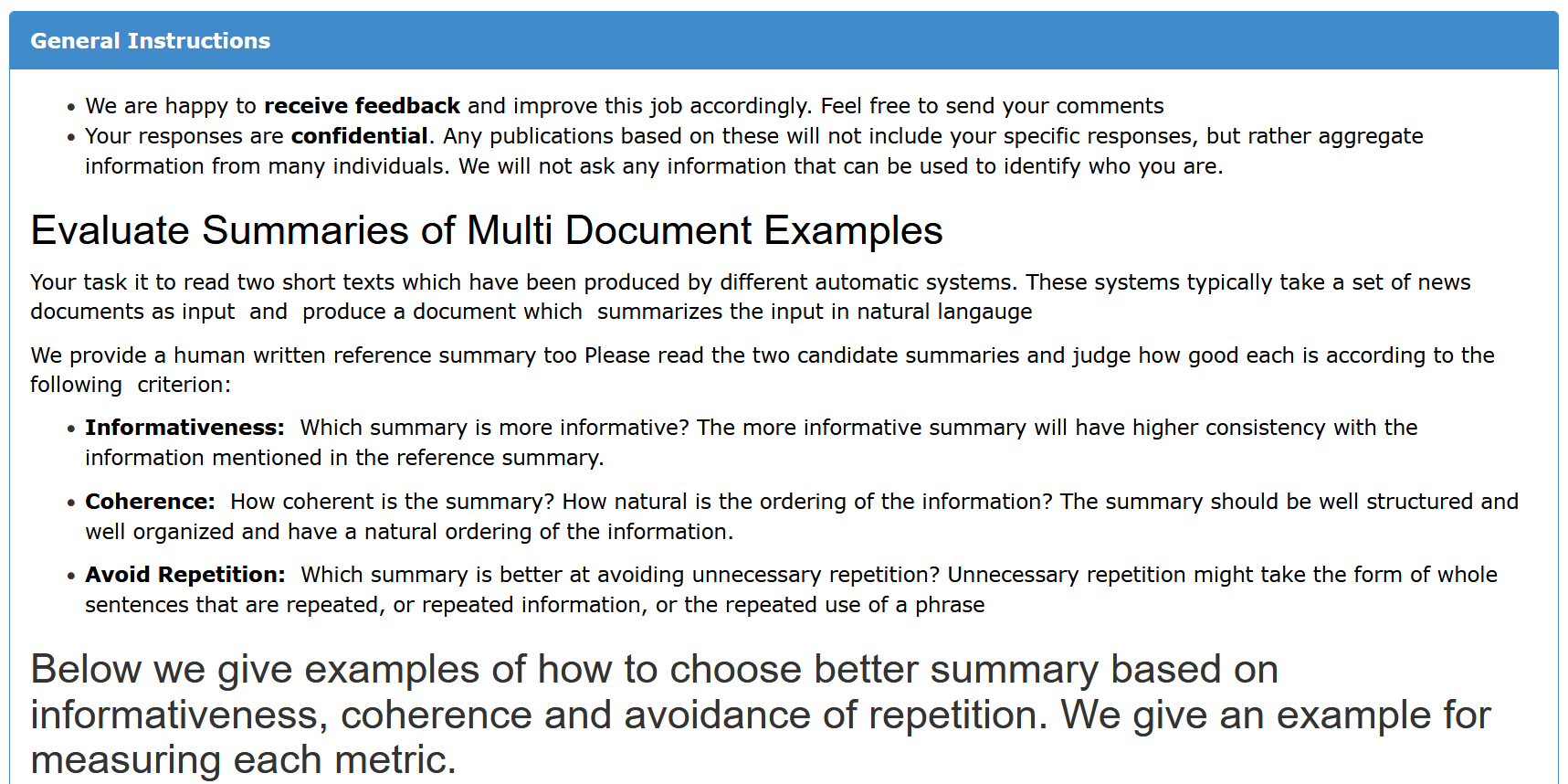}
    \includegraphics[width=0.9\textwidth]{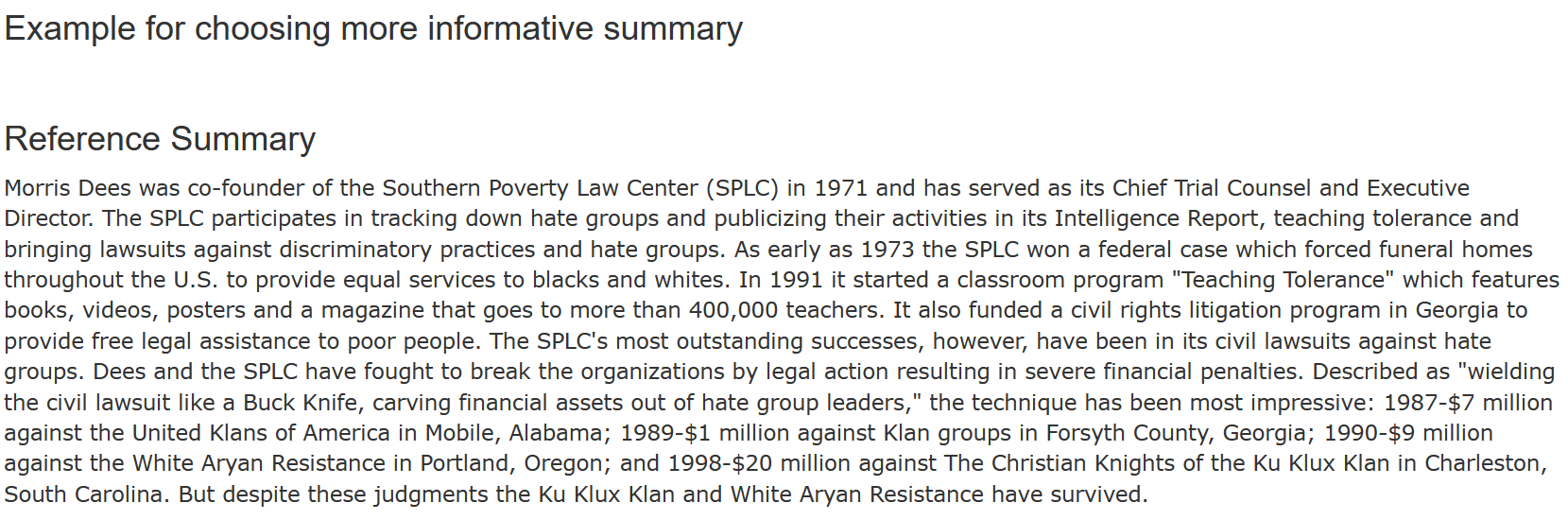}
    \includegraphics[width=0.9\textwidth]{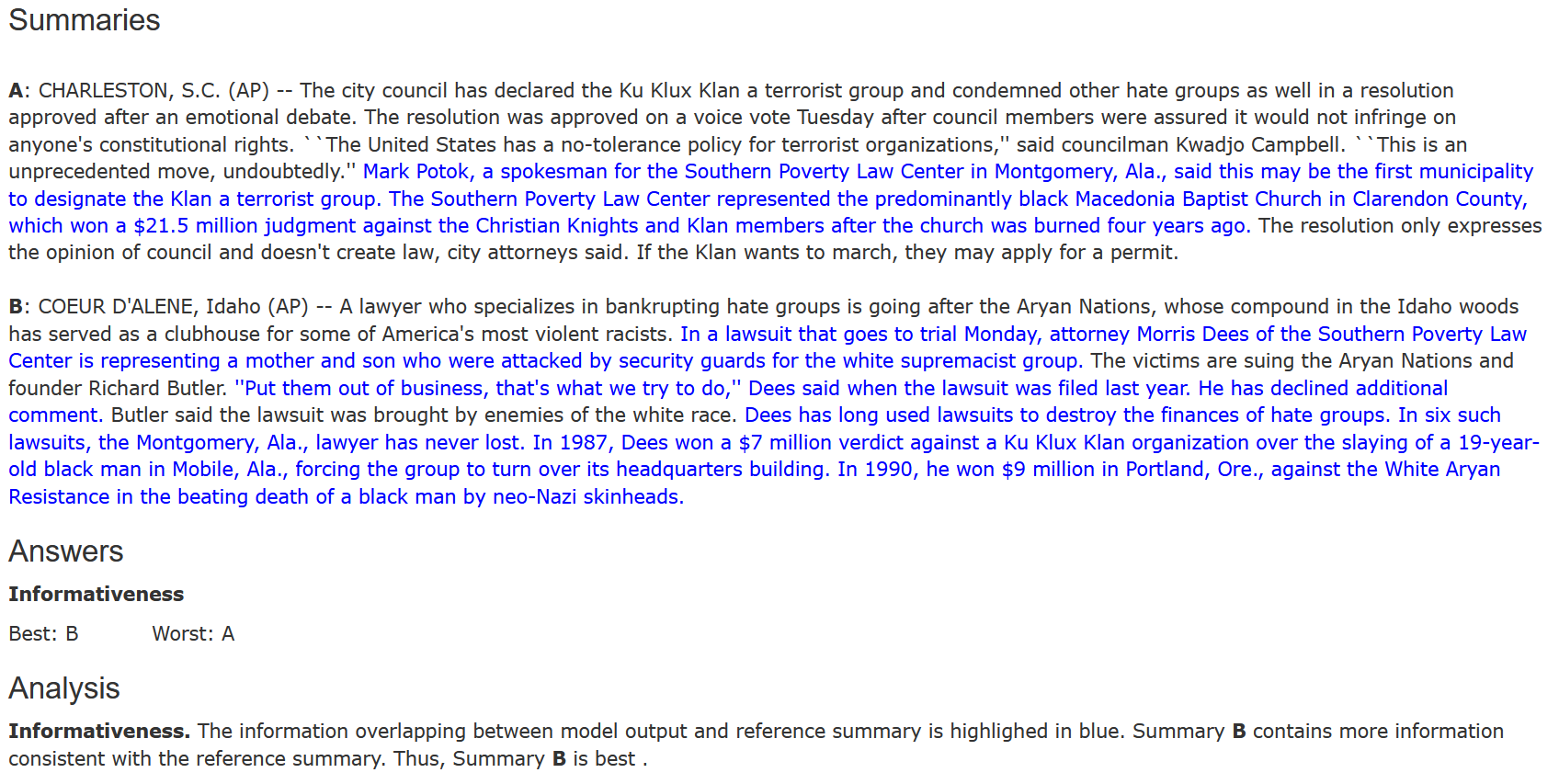}    
    \caption{Instructions for human evaluation}
    \label{fig:my_label}
\end{figure*}

\begin{figure*}
    \centering
    \includegraphics[width=0.9\textwidth]{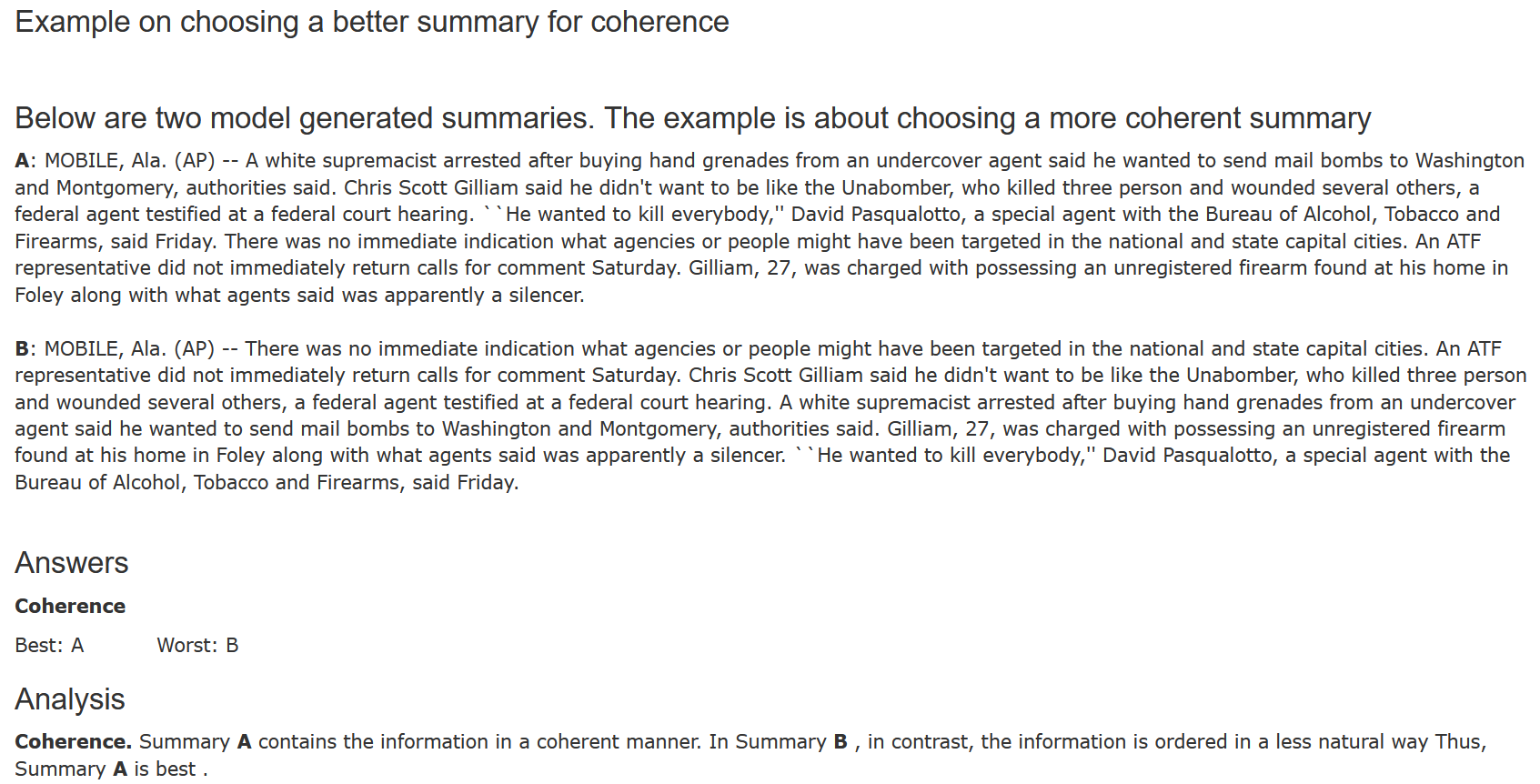}
    \includegraphics[width=0.9\textwidth]{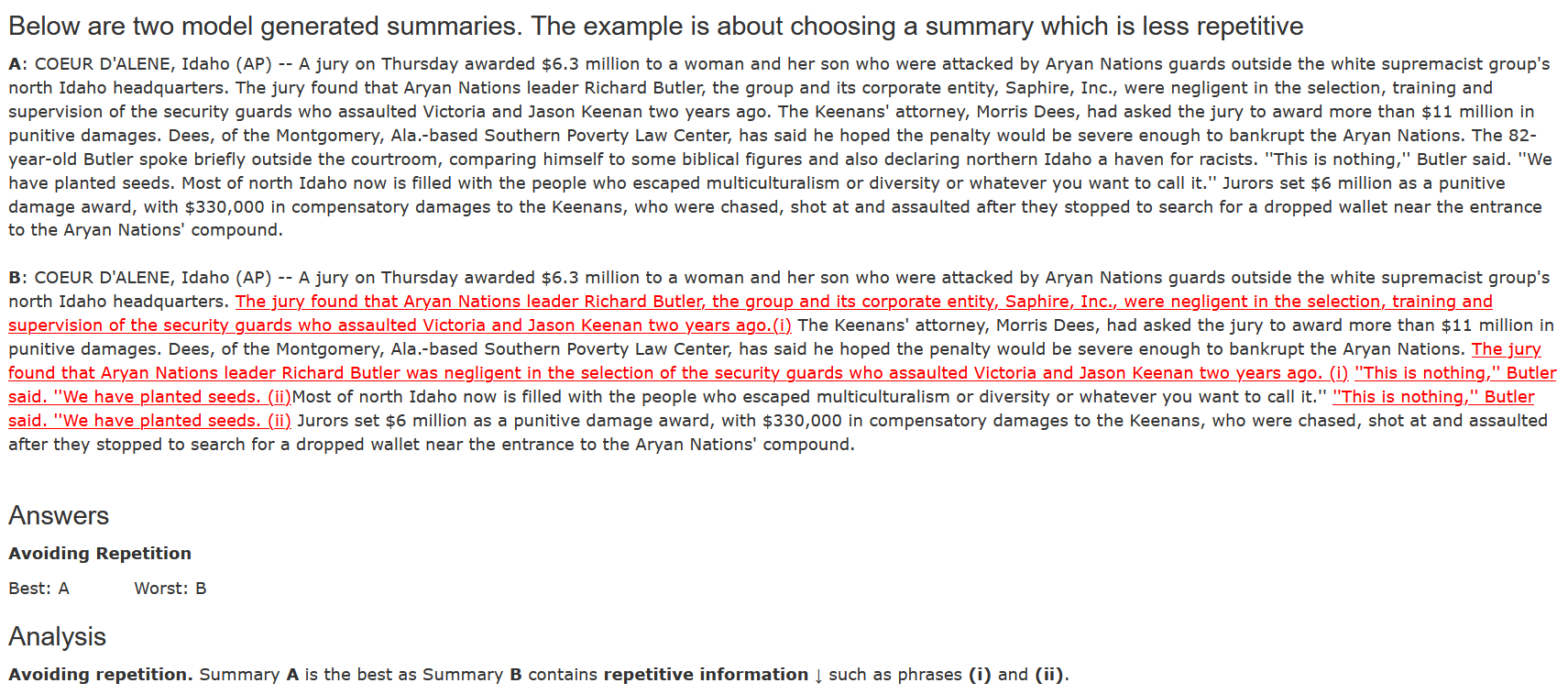}    
    \caption{Instructions for human evaluation (continued)}
    \label{fig:my_label2}
\end{figure*}

Figures \ref{fig:my_label} and \ref{fig:my_label2} show the screenshots of the user interface presented to the raters.
\end{document}